\PassOptionsToPackage{table}{xcolor}
\documentclass{vllm}
\usepackage[margin=1.2in]{geometry}
\usepackage{amsmath,amssymb,tabularx}
\usepackage[numbers,sort&compress]{natbib}
\usepackage{enumitem,flafter}
\setlist{nosep,leftmargin=*}
\newcommand{\LN}{\operatorname{LN}}
\newcommand{\release}[2]{\href{https://huggingface.co/llm-semantic-router/Decision-1.0-#1/tree/#2}{#1}}
\hypersetup{pdftitle={Decision 1.0: Towards Open Decision Foundation Models},pdfauthor={Xunzhuo Liu, Haichen Zhang, Andy Luo, Huamin Chen, Xue Liu, Bowei He}}
\title{Decision 1.0}
\subtitle{Towards Open Decision Foundation Models}
\author{Xunzhuo Liu$^{1,2}$, Haichen Zhang$^{1,2}$, Andy Luo$^{1,2}$,\\Huamin Chen$^{1,3}$, Xue Liu$^{1,4,5}$, Bowei He$^{1,4,5,\dagger}$ \\ $^1$vLLM Semantic Router Project, $^2$AMD, $^3$Microsoft, $^4$MBZUAI, $^5$McGill University \\ $\dagger$ Corresponding Author: \texttt{Bowei.He@mbzuai.ac.ae}}
\date{September 2026}
\abstract{Intelligent software needs reusable judgments inside its control flow: selecting an action, testing a condition, or assessing evidence against a rubric. We present Decision 1.0, an open-weight family that turns textual state and runtime-defined questions into typed probability distributions. Its Choice, Noul, and Score interface lets applications specify outcomes at inference time and compose predictions into agents and workflows. Each question produces its candidate distribution without autoregressive text generation. Six models cover bidirectional encoder and causal decoder architectures, including a workflow specialist. Training draws on verifiable tasks, natural-language annotations, and workflow adaptation. On a common decision benchmark, Lux-9B achieves 84.17\% on explicit decisions and 91.25\% on inference, with a weighted overall score of 77.22\%. Results also expose gaps in composition, transfer, and probability calibration. Released weights and inference code provide a basis for developing decision models as reusable components of software.}
\begin{document}
\maketitle

\section{Introduction}
As learned models become components of software, their outputs must support decisions inside programs. Routing, agent control, and workflow automation require judgments that can be combined with explicit state, constraints, and rules. A reusable model for this setting should interpret a task supplied at runtime and return probabilities over the application's allowed outcomes.

Jev's System One interface highlights this opportunity through typed, probabilistic decisions~\cite{jev}. Developing the idea as an open research direction requires access to model weights and implementations: researchers need to examine how representations and training affect decisions, adapt models to new domains, and test whether their probabilities remain useful as tasks change.

We introduce Decision 1.0, six open-weight models spanning bidirectional encoders and causal decoders under a common Choice, Noul, and Score interface. The family includes general decision models and a workflow specialist, with documented architectures, training sources, and evaluation. Our aim is to establish an open foundation for programmable decision-making: models that can be adapted to different applications while preserving an interface that software can compose and inspect. This release studies the current capabilities of that approach and the remaining challenges in composition, transfer, and calibration.

\section{Task Formulation}
A request contains textual state $x$ and questions $\{q_j\}_{j=1}^{Q}$. A question specifies a type $t$, instructions $u$, and candidate descriptions $C=(c_1,\ldots,c_K)$. The model predicts
\begin{equation}
 p_\theta(i\mid x,u,C,t)=\frac{\exp(z_i/T)}{\sum_{k=1}^{K}\exp(z_k/T)},
 \label{eq:distribution}
\end{equation}
where $z_i$ is a candidate score and $T>0$ is an inference temperature. Padded candidates are excluded from normalization.

\textbf{Choice} returns this distribution and the identifier of its maximum. \textbf{Noul}, the interface name for a Boolean judgment, returns $p(\mathrm{true})$ over the alternatives false and true. \textbf{Score} returns a distribution over ordered rubric levels and its expected index, $\sum_i(i-1)p_i$. The native encoder interface also supports expectations over supplied numeric values.

Candidate descriptions define each task at runtime. Applications can include an explicit abstention or unknown outcome when the available evidence may not support a decision.

The complete input includes state, instructions, candidates, and special tokens. The public interfaces accept up to 1,024 tokens for encoders and 16,384 for causal models, rejecting overlength requests. Choice supports 2--255 candidates and the common Score interface supports 2--10 levels. Questions are computed independently and may be batched.

\section{Model Architectures}
\label{sec:architecture}
Decision 1.0 combines two architectures under one typed interface (Table~\ref{tab:models}). Kai and Lex use bidirectional encoders; Eos, Sol, Nox, and Lux use causal backbones. Both score candidates supplied at runtime, but differ in how they encode the evidence and extract candidate representations. Each checkpoint has independent weights.

\begin{table}[!htbp]
\centering\small
\caption{Decision 1.0 model family. Parameter counts include the deployed decision readout; context covers the complete packed question. Lex is evaluated separately as a workflow specialist.}
\label{tab:models}
\setlength{\tabcolsep}{4pt}
\begin{tabularx}{\textwidth}{llrrX}
\toprule
Model & Initialization & Parameters & Context & Architecture / scope \\
\midrule
Kai-0.6B & Vela / mmBERT & 0.57B & 1K & Bidirectional; general \\
Lex-0.6B & Earlier Kai & 0.57B & 1K & Bidirectional; workflow \\
Eos-0.8B & Qwen3.5-0.8B & 0.75B & 16K & Causal; general \\
Sol-2B & Qwen3.5-2B & 1.88B & 16K & Causal; general \\
Nox-4B & Qwen3.5-4B & 4.21B & 16K & Causal; general \\
Lux-9B & Qwen3.5-9B & 7.94B & 16K & Causal; general \\
\bottomrule
\end{tabularx}
\end{table}

\subsection{Encoder-based implementation: Kai and Lex}
\begin{figure}[!htbp]
\centering
\begin{minipage}[t]{0.475\textwidth}
\setlength{\parskip}{5pt}
\vspace{0pt}
Kai is initialized from Vela Encoder~\cite{vela}, a derivative of mmBERT~\cite{mmbert} using the ModernBERT architecture~\cite{modernbert}. The packed sequence places the typed question first, then each candidate description preceded by a \texttt{[MASK]} marker, followed by the state. Joint bidirectional attention gives a marker access to its candidate, the evidence, and competing outcomes. External candidate IDs are bookkeeping rather than input tokens.

Within a checkpoint, token embeddings, embedding normalization, and a three-entry type embedding table are shared. Choice, Noul, and Score each have their own 22-layer backbone path, final LayerNorm, two interaction layers, and scalar scorer. \end{minipage}\hfill
\begin{minipage}[t]{0.49\textwidth}
\vspace{0pt}
\centering
\includegraphics[width=\linewidth]{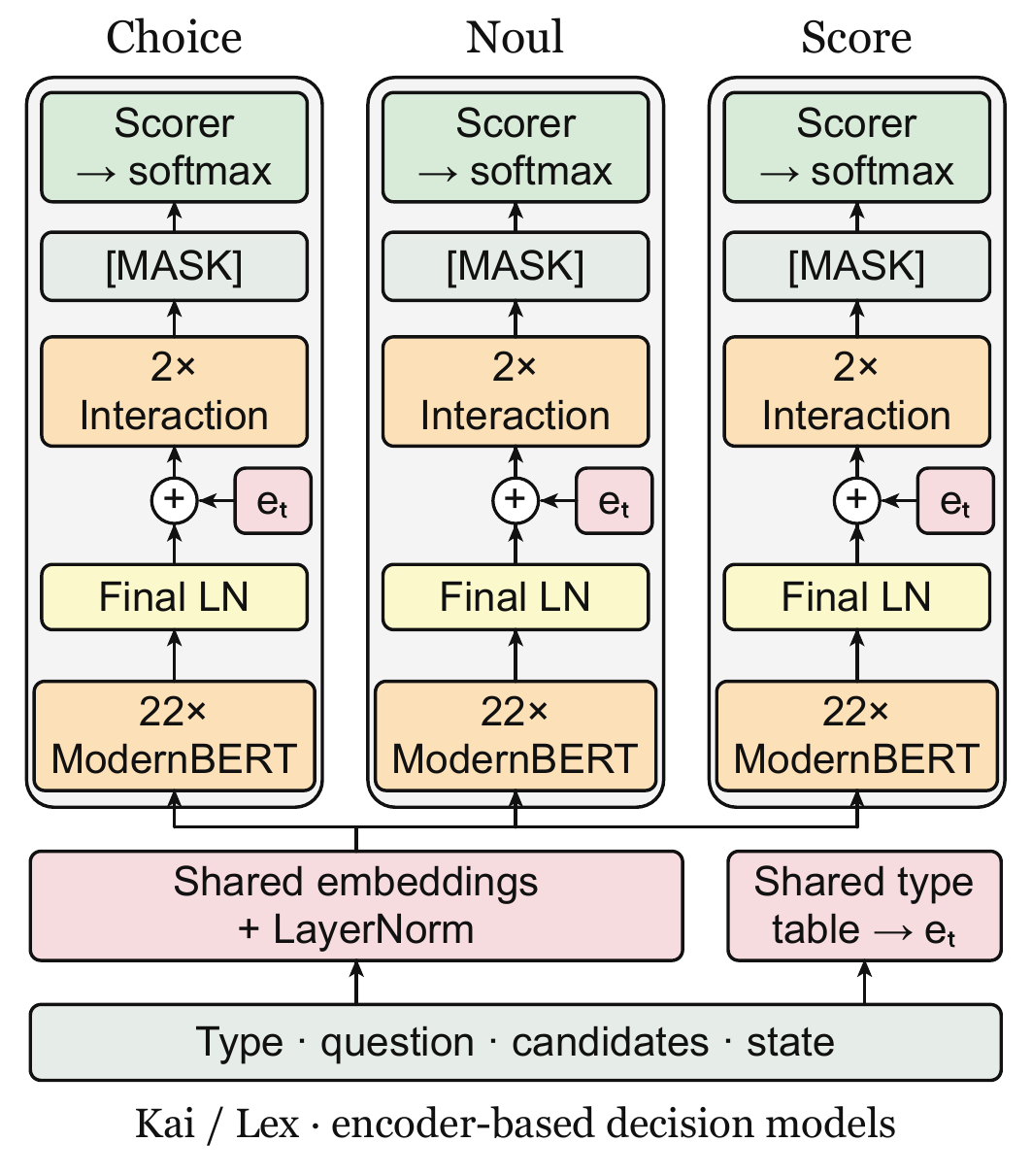}
\end{minipage}
\captionsetup{font=footnotesize}
\caption{Kai and Lex: shared embeddings feed three independent type paths. Each path has its own encoder, interaction layers, and scorer. $e_t$ comes from the shared type table.}
\label{fig:encoder}
\end{figure}

\paragraph{Inputs and type paths.}
The encoder jointly represents the question, candidate descriptions, and state. A shared embedding layer feeds independent paths for Choice, Noul, and Score (Figure~\ref{fig:encoder}).

\begin{figure}[!htbp]
\centering
\includegraphics[width=\textwidth,height=0.45\textheight,keepaspectratio]{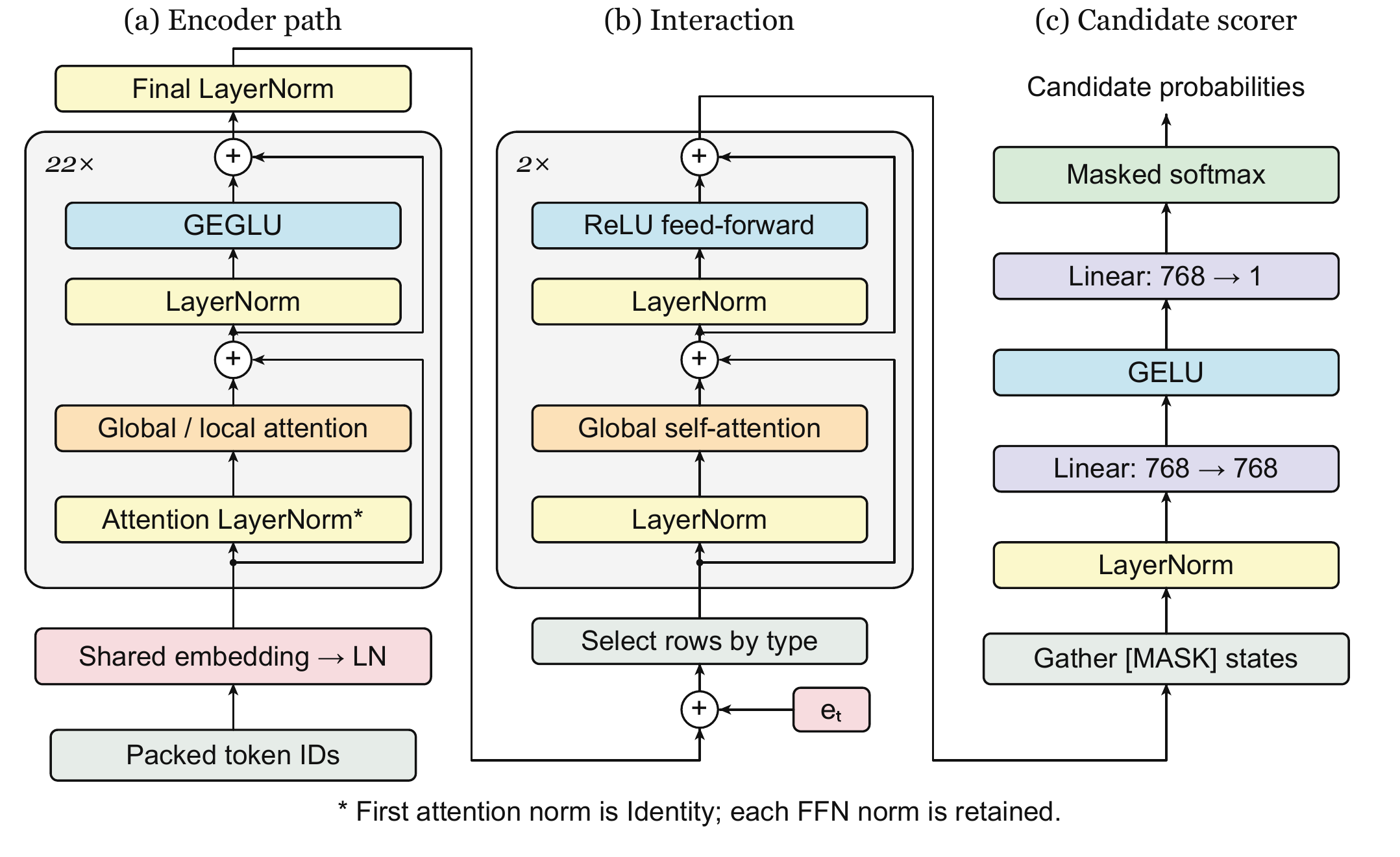}
\caption{Expanded encoder computation: a ModernBERT backbone, two type-conditioned interaction layers, and a candidate scorer. The first backbone attention normalization is Identity.}
\label{fig:encoder-path}
\end{figure}

\paragraph{Backbone and interaction.}
Figure~\ref{fig:encoder-path} expands a single type path. The backbone has width 768 and pre-LayerNorm residual blocks with GEGLU feed-forward width 1,152. The attention normalization in the first layer is Identity because the embeddings have already been normalized. Global attention appears at zero-based layers $0,3,\ldots,21$; the remaining layers use a bidirectional local window of $\pm64$ tokens. Rotary position transformations act on queries and keys.

After path-specific final normalization, the selected type embedding is added to every token. Two additional pre-LayerNorm interaction layers use global self-attention and a ReLU feed-forward network of width 3,072. 

\begin{figure}[!htbp]
\centering
\includegraphics[width=\textwidth,height=0.45\textheight,keepaspectratio]{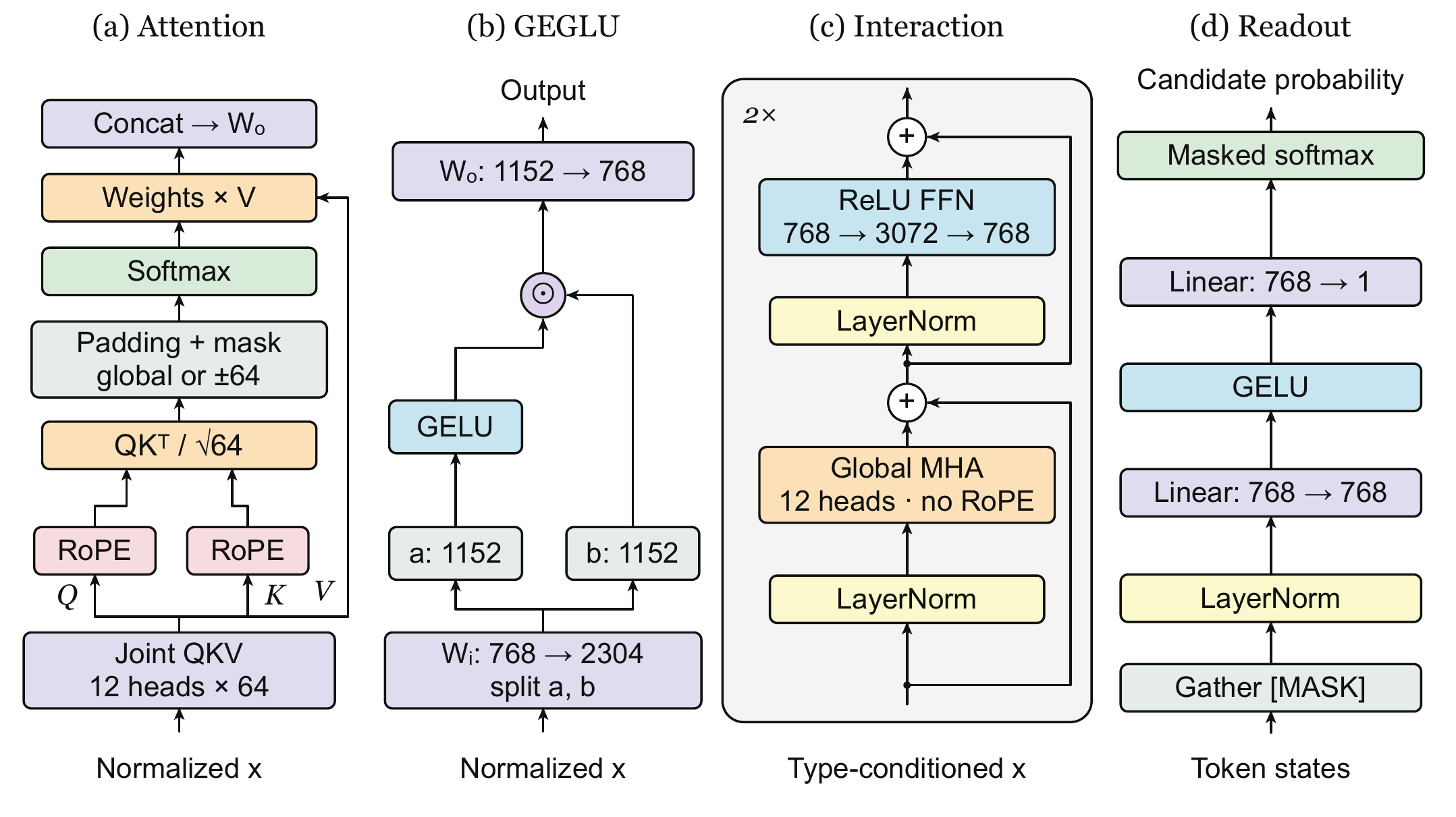}
\caption{Encoder operators: rotary attention, GEGLU, the pre-norm interaction layer, and the type-specific scorer. Dropout is inactive at inference.}
\label{fig:encoder-operators}
\end{figure}

\paragraph{Candidate scoring.}
Following Laya's candidate-marker design~\cite{laya}, the model gathers the contextualized marker $h_i$ for each candidate and applies
\begin{equation}
 z_i=w_t^\top\operatorname{GELU}(W_t\LN_t(h_i)+b_t)+a_t,
 \qquad W_t\in\mathbb{R}^{768\times768}.
 \label{eq:encoderhead}
\end{equation}
Figure~\ref{fig:encoder-operators} shows the attention, feed-forward, and readout operators. The scorer is shared across candidates within each type. Lex retains the same topology with independently adapted weights. In native mixed-type batches, each active backbone path processes the full padded batch; rows are selected by type before the interaction layers.

\subsection{Decoder-based implementation: Eos, Sol, Nox, and Lux}
\begin{figure}[!htbp]
\centering
\begin{minipage}[t]{0.475\textwidth}
\setlength{\parskip}{5pt}
\vspace{0pt}
The decoder-based models inherit the Qwen3.5 text-backbone topology~\cite{qwen35}. Eos and Sol have 24 layers, while Nox and Lux have 32. Each group of four layers uses three Gated DeltaNet mixers followed by one full causal attention layer. Both layer types have two pre-RMSNorm residual sublayers: token mixing and a SwiGLU feed-forward network. A separate RMSNorm follows the complete layer stack.

The input orders state, typed question, candidate blocks, and a final decision suffix. The question type is represented in the token sequence, and each candidate block includes its key and description.

The backbone uses BF16; decision scoring uses FP32. The vision tower and vocabulary-generation head are absent from decision inference. \end{minipage}\hfill
\begin{minipage}[t]{0.49\textwidth}
\vspace{0pt}
\centering
\includegraphics[width=\linewidth]{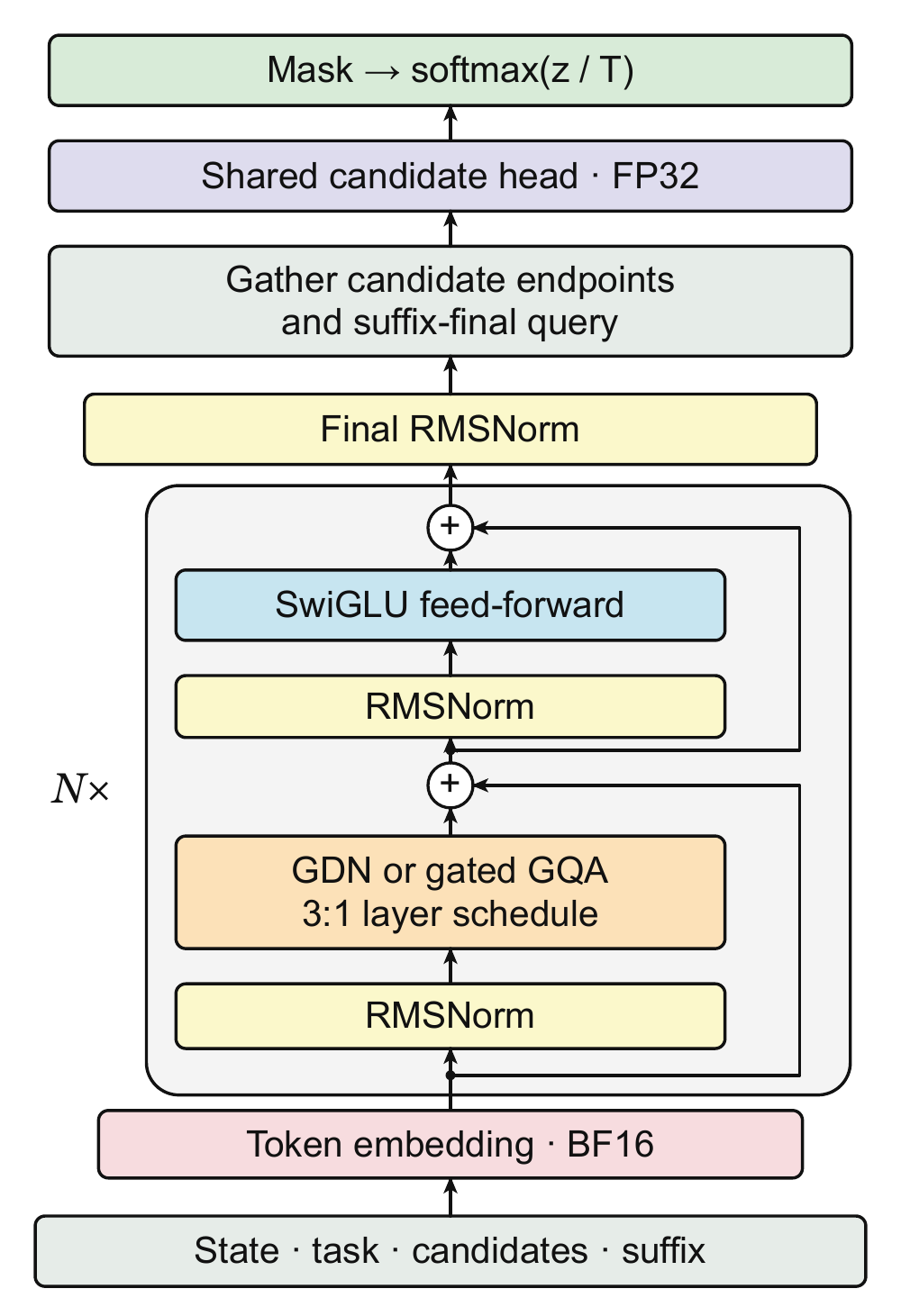}
\end{minipage}
\captionsetup{font=footnotesize}
\caption{Eos, Sol, Nox, and Lux: a hybrid causal backbone feeds one shared FP32 candidate readout. $N=24$ for Eos/Sol and $N=32$ for Nox/Lux.}
\label{fig:causal}
\end{figure}

\paragraph{Backbone and input packing.}
The causal family inherits a Qwen3.5 text backbone and replaces vocabulary prediction with candidate scoring. Figure~\ref{fig:causal} shows the complete inference path.

\FloatBarrier
\paragraph{Candidate-conditioned readout.}
The readout combines each candidate endpoint with a query that summarizes the complete input. Let $H$ be the final-normalized sequence, $e_i$ the last token of candidate block $i$, and $L$ the unpadded sequence length. The readout gathers $h_i=H[e_i]$ and $h_q=H[L-1]$, using the suffix-final state to condition every candidate score on the complete input.

\begin{figure}[!htbp]
\centering
\includegraphics[width=\textwidth,height=0.45\textheight,keepaspectratio]{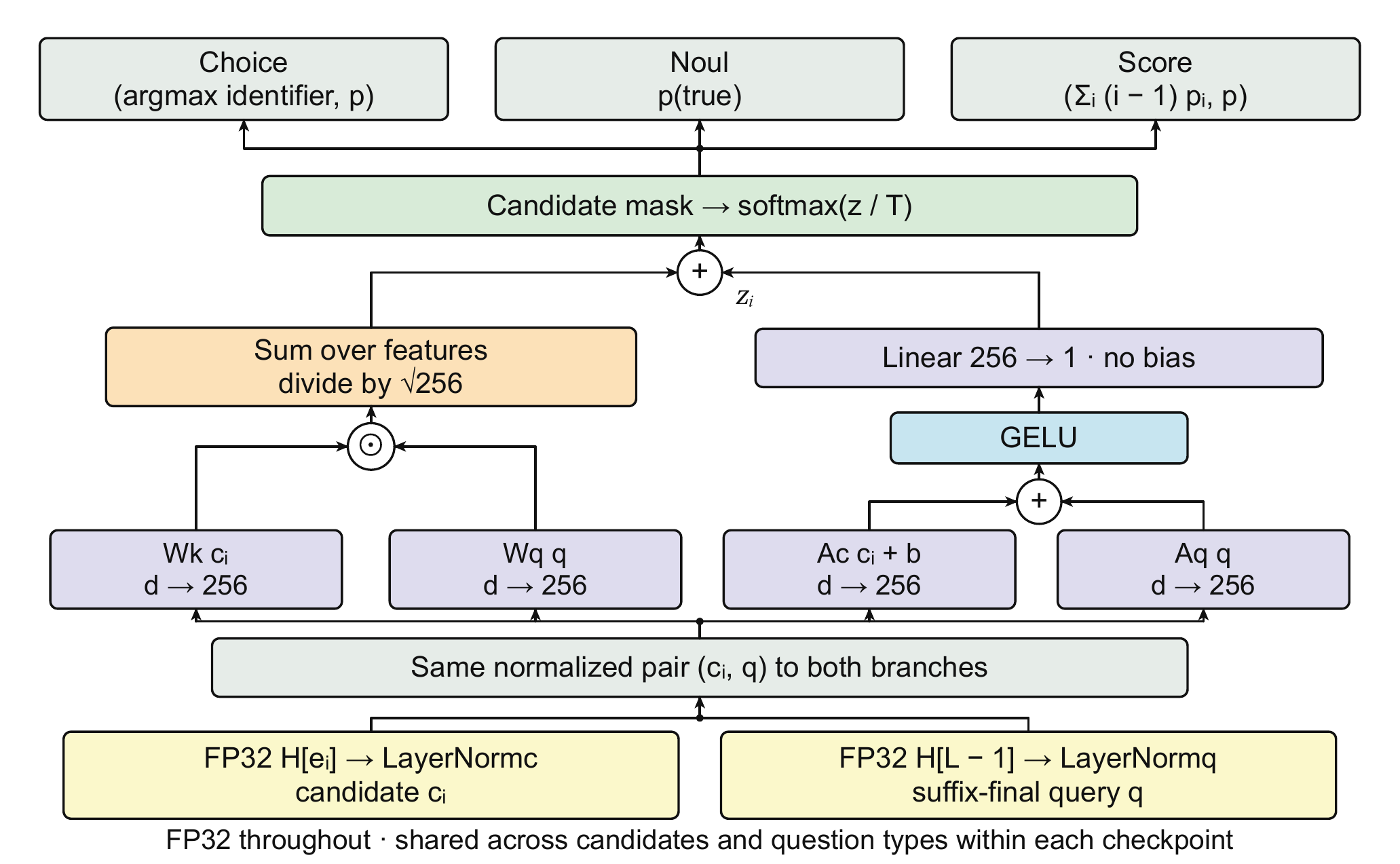}
\caption{FP32 candidate readout (Equation~\ref{eq:head}). Four independent projections form bilinear and nonlinear scores; candidate masking and temperature scaling precede typed output conversion.}
\label{fig:decoder-readout}
\end{figure}

Separate LayerNorms yield $c_i=\LN_c(h_i)$ and $q=\LN_q(h_q)$. The shared FP32 head adds a bilinear compatibility score and a nonlinear interaction:
\begin{equation}
 z_i=\frac{(W_kc_i)^\top(W_qq)}{\sqrt{d_h}}
      +w^\top\operatorname{GELU}(A_cc_i+A_qq+b),\qquad d_h=256.
 \label{eq:head}
\end{equation}
The readout uses four independently parameterized projections; only the candidate-side nonlinear projection carries a bias. Within one checkpoint, this head is shared across all candidates and all three decision types. Padded candidates are masked before temperature scaling and softmax. Figure~\ref{fig:decoder-readout} makes the two score components explicit.

An early candidate endpoint cannot attend to later candidates, but the final query conditions every score on the complete choice set. This supports variable output cardinality without enforcing permutation invariance.

Both architectures return all candidate probabilities after one complete forward pass per question. The current implementation does not reuse state activations across questions; eliminating output-token generation alone does not establish a latency advantage over a one-token generative classifier.

\section{Data and Training}
\label{sec:training}
The central training problem is to bind a textual task definition to its allowed outcomes while preserving useful language understanding. Procedural supervision supplies explicit rules, controlled state changes, and verifiable answers. Natural-language annotations provide reading, intent, inference, and rubric judgments that templates alone do not cover. The documented Sol, Nox, and Lux mixtures do not use official Jev predictions as training labels.

\subsection{Distribution supervision}
In the documented Lex, Sol, Nox, and Lux distribution-supervised stages, a target $y\in\Delta^{K-1}$ gives the objective
\begin{equation}
 \mathcal{L}_{\mathrm{CE}}=-\sum_{i=1}^{K}y_i\log p_\theta(i\mid x,u,C,t),
\end{equation}
with $T=1$ during training. Hard labels become one-hot distributions; source soft labels retain their recorded distribution. Human ordinal means can be represented by interpolation between adjacent levels, preserving the annotated mean. Candidate reordering must also reorder its targets, and ordered rubric levels cannot be shuffled without preserving their semantics.

\subsection{Encoder adaptation and workflow specialization}
Kai inherits multilingual supervision spanning inference, intent, evidence, and human ratings. Representative sources include SNLI, MASSIVE, BoolQ, TyDi QA, HelpSteer2, and OpenAssistant~\cite{snli,massive,boolq,tydi,helpsteer,oasst}. Their annotations are converted into candidate decisions with supporting context where required.

Kai adapts Choice and Score paths on approximately 21,000 complete-input examples with frozen shared embeddings, retaining an earlier Noul path in the released checkpoint. Lex specializes an earlier Kai checkpoint on 6,000 generated training decisions from \texttt{typed-decisions}~\cite{typeddata}, covering customer service, invoices, security incidents, and agent traces. It combines teacher-derived soft targets and hard labels equally, adapts all three paths with frozen embeddings, and follows a fixed refit schedule before evaluation on the 2,000-decision test set.

\subsection{Causal adaptation and retention}
Sol and Nox use an English/Chinese curriculum of procedural decisions, complemented by intent and inference annotations~\cite{banking,clinc,mnli}. Replay retains earlier task coverage. Natural-language adaptation uses a 24,000-example mixture of retained decisions, Cosmos QA, SQuAD 2.0 answerability, and SNLI~\cite{cosmos,squad,snli}. Examples from each source-parent or detected near-duplicate group are assigned to the same training, selection, or calibration split. Known evaluation matches are excluded from custom training construction.

Sol uses candidate cross-entropy training. Nox adds supervised retention through frozen vocabulary rows for legal answer letters, with this auxiliary readout removed at deployment. Lux adapts Qwen3.5-9B on a 24,000-example mixture of retained decisions and human-annotated judgments. Its released final stage uses hard-label cross-entropy and replaces eligible opaque option keys with semantic descriptions while preserving the evidence, candidate descriptions, order, and labels.

Eos uses head-only warmup followed by full-model adaptation over 24,000 training rows. Differences in initialization and training recipes make the family a comparison of released systems rather than a controlled scaling or attention-direction study.

\subsection{Selection and calibration}
Causal checkpoint selection emphasizes decision accuracy, with probability quality used to break ties where documented. A positive scalar temperature is fitted separately by minimizing negative log likelihood~\cite{guo}, adjusting distribution sharpness while preserving the argmax. Lex uses a fixed final schedule and no test-based temperature fitting.

\section{Evaluation}
\label{sec:eval}
We evaluate general decision quality, workflow specialization, and probability quality separately. The common benchmark compares five general Decision models with ten reference systems; Lex is assessed on its dedicated workflow test. All results use the fixed public revisions in Appendix~\ref{app:releases}.

\subsection{Setup and protocol}
\label{sec:protocol}
The common benchmark contains 3,766 scored decisions across 54 task slices. Five panels measure different capabilities: \emph{Decisions} covers explicit classification, Boolean rules, and rubric judgments; \emph{Composition} tests interacting constraints and state updates; \emph{Reading} tests comprehension of provided evidence; \emph{Inference} tests answerability, entailment, and commonsense judgments; and \emph{Transfer} tests reuse on the external Kev suite~\cite{kev}. The suite covers English and Chinese. Appendix~\ref{app:eval-details} gives the source datasets, sample counts, and aggregation rules.

Systems receive the same examples and candidate order through their native inference interfaces. We compare the official Jev 1.13.0 service, English and multilingual Laya, Decider, three Kev sizes, and untuned Qwen3.5 references~\cite{jev,laya,decider,kev,qwen35}. Panel scores measure categorical accuracy, including for Score questions. Because much of the benchmark was observed repeatedly during development, these results characterize the released systems rather than establish performance on a fresh blind test.

\subsection{General decision accuracy}
Lux leads Decisions and Inference among the 15 systems in Table~\ref{tab:general}, at 84.17\% and 91.25\%, and ranks second by the published Overall score. Relative to its untuned Qwen3.5-9B reference, these are gains of 10.26 and 11.67 percentage points. Sol and Nox likewise improve Decisions by 16.62 and 13.11 points over their corresponding references. Eos exceeds Kev-0.8B on the four non-transfer panels, and Kai exceeds both general Laya models on those panels. These system-level comparisons do not isolate the effects of the readout, data, and optimization recipe.

\begin{table}[!htbp]
\centering
\caption{General decision accuracy (\%). Bold marks the best result in each column; our models are shaded and marked (ours). Rank follows the published, post-hoc weighted Overall score; aggregation is specified in Appendix~\ref{app:eval-details}.}
\label{tab:general}
\setlength{\tabcolsep}{3pt}
\renewcommand{\arraystretch}{1.14}
\fontsize{8}{9.5}\selectfont
\begin{tabularx}{\textwidth}{@{}r>{\raggedright\arraybackslash}Xrrrrrr@{}}
\toprule
Rank & Model & Overall & Decisions & Composition & Reading & Inference & Transfer \\
\midrule
1 & Jev 1.13.0 & \textbf{81.05} & 79.10 & \textbf{66.38} & \textbf{94.53} & 89.79 & \textbf{87.19} \\
\rowcolor{metabg} 2 & \textbf{Lux-9B (ours)} & 77.22 & \textbf{84.17} & 52.75 & 89.69 & \textbf{91.25} & 77.63 \\
\rowcolor{metabg} 3 & \textbf{Nox-4B (ours)} & 73.09 & 83.00 & 51.79 & 79.06 & 86.25 & 69.60 \\
4 & Kev-9B & 71.89 & 76.75 & 45.75 & 86.72 & 83.54 & 79.25 \\
5 & Kev-4B & 70.09 & 71.90 & 48.54 & 81.88 & 84.58 & 76.10 \\
6 & Qwen3.5-9B & 69.73 & 73.91 & 44.62 & 89.84 & 79.58 & 73.23 \\
7 & Decider 2B & 67.71 & 64.01 & 46.58 & 92.03 & 84.38 & 69.31 \\
8 & Qwen3.5-4B & 67.29 & 69.89 & 43.33 & 87.97 & 79.79 & 68.83 \\
\rowcolor{metabg} 9 & \textbf{Sol-2B (ours)} & 66.32 & 73.75 & 46.08 & 76.56 & 84.17 & 57.07 \\
\rowcolor{metabg} 10 & \textbf{Eos-0.8B (ours)} & 61.89 & 65.94 & 46.04 & 70.31 & 81.67 & 52.01 \\
11 & Kev-0.8B & 58.28 & 60.14 & 42.29 & 67.81 & 68.75 & 61.19 \\
12 & Qwen3.5-2B & 57.24 & 57.12 & 39.00 & 73.75 & 72.29 & 56.31 \\
\rowcolor{metabg} 13 & \textbf{Kai-0.6B (ours)} & 53.52 & 57.96 & 40.83 & 54.69 & 69.79 & 48.37 \\
14 & Laya English & 51.03 & 56.54 & 35.33 & 51.41 & 63.75 & 53.06 \\
15 & Laya Multilingual & 47.19 & 47.25 & 38.92 & 50.78 & 57.29 & 47.13 \\
\bottomrule
\end{tabularx}
\end{table}

The gains do not extend uniformly across capabilities. Jev leads Overall, Composition, Reading, and Transfer; Lux remains 13.62 points behind it on Composition and 9.56 on Transfer. Kev-9B also exceeds Lux on Transfer. Adaptation can trade off existing capabilities: Nox's reading accuracy falls 8.91 points below its untuned reference. Transfer also remains weaker than Kev-0.8B for Eos and Laya English for Kai. Composition and transfer therefore remain priorities for broader decision reuse.

\subsection{Workflow specialization}
On the separate \texttt{typed-decisions} test of 2,000 judgments from 400 state groups, Lex reaches 78.15\% accuracy versus 76.60\% for the official Laya specialist. The paired group-bootstrap 95\% interval for this 1.55-point difference is $[-0.20,+3.15]$; the overall advantage is not statistically established. Lex has higher Score accuracy, whereas Noul accuracy and probability-quality measures favor Laya. Specialization yields a mixed profile rather than a uniform improvement.

\subsection{Probability quality}
A useful decision distribution must convey more than its most likely label. Table~\ref{tab:calibration} compares Lux with the two systems that achieve higher transfer accuracy. Lux is close to Kev-9B in Brier score and negative log likelihood, but has higher calibration error. Among these three systems, Jev leads all three measures. Transfer calibration remains a limitation despite temperature fitting.

\begin{table}[!htbp]
\centering\small
\caption{Probability quality for Lux and its higher-accuracy transfer comparators. Lower is better; bold marks the best value among the displayed systems. ECE is reported in percent.}
\label{tab:calibration}
\begin{tabular}{lrrr}
\toprule
Model & Brier & NLL & ECE (\%) \\
\midrule
Jev 1.13.0 & \textbf{0.191} & \textbf{0.574} & \textbf{3.35} \\
\rowcolor{metabg} Lux-9B (ours) & 0.294 & 0.594 & 5.56 \\
Kev-9B & 0.295 & 0.603 & 4.45 \\
\bottomrule
\end{tabular}
\end{table}

Candidate order and missing evidence present separate robustness challenges. Neither architecture enforces candidate-order invariance, and a closed candidate set cannot express an abstention outcome that the application has not supplied.

\section{Related Work}
Candidate-conditioned classification connects to entailment-based zero-shot classification, where labels are supplied as descriptions rather than fixed output indices~\cite{yin}. ModernBERT and mmBERT provide bidirectional representations; Qwen3.5 provides the causal initialization used here~\cite{modernbert,mmbert,qwen35}. The decision interface is compatible with either attention direction.

Jev, Laya, Decider, and Kev explore typed probabilistic interfaces and candidate readouts~\cite{jev,laya,decider,kev}. Decision builds on these designs with a released encoder and causal family, multilingual and workflow adaptation, and a common evaluation of accuracy, transfer, and probability quality. Temperature scaling addresses calibration separately from classification accuracy~\cite{guo}.

\section{Future Directions}
Future work targets the cost of multi-question inference while preserving decision quality.

\paragraph{Reuse state computation and improve batching.}
An identical causal prefix could be computed once and branched across questions, reusing attention key/value caches together with Gated DeltaNet recurrent and convolutional states. Length-aware batching and serving methods such as FlashInfer~\cite{flashinfer} motivate further study on this hybrid backbone. Evaluation should vary input length, candidate count, question count, and batch size, measuring throughput and tail latency.

\paragraph{Compress the model while preserving decisions.}
Quantization methods such as SmoothQuant~\cite{smoothquant} motivate lower-precision backbone execution with a higher-precision readout. Distillation of candidate distributions, including methods such as decoupled knowledge distillation~\cite{dkd}, could transfer a larger model's behavior to a smaller decision model. Accuracy, calibration, and candidate-order robustness should guide these studies.

\paragraph{Learn a reusable state representation.}
Perceiver IO and in-context autoencoding~\cite{perceiver,icae} motivate a compact state representation shared by question-specific readers. This would require a new trained architecture that preserves evidence needed by questions beyond those used to construct the representation.

\paragraph{Allocate computation to difficult questions.}
Routing and cascading can allocate model capacity under a cost--quality objective~\cite{routingcascade}. Within-model early exits would additionally require trained intermediate heads and validated exit confidence. Policies for grouped requests should be assessed on task-held-out data using per-question quality and end-to-end latency.

\section{Limitations and Availability}
Decision 1.0 is text-only and relies on supplied evidence. The general evaluation covers English and Chinese; quality depends on the task, input budget, evidence, and candidate set. Training and evaluation reuse public task families, and upstream pretraining exposure is unknown. Differences in recipes and native input handling limit attribution to individual architectural changes.

We release weights, inference implementations, configurations, and evaluation documentation~\cite{decision}; encoder releases also include continued fine-tuning code. Not every assembled training corpus, construction script, or final optimization setting is distributed. Project artifacts use Apache 2.0 with retained upstream notices; source datasets retain their own terms.

\section{Conclusion}
Decision 1.0 maps runtime-defined alternatives directly to typed probabilities through bidirectional or causal models. The family improves explicit decision performance and supports workflow adaptation, while composition, transfer, order sensitivity, and uncertainty remain open problems. Released weights and inference code provide a basis for further research.

\section*{Acknowledgments}
We thank the authors and maintainers of the pretrained models, datasets, and open decision implementations on which this work builds.

\setlength{\bibsep}{1.5pt}

\appendix
\section{Released Artifacts}
\label{app:releases}
The results describe fixed public model snapshots. The following links resolve to the exact revisions used in this report:
\release{Kai-0.6B}{7185f514f54b8f93c55998b1e8f9c5cc67f0d029},
\release{Lex-0.6B}{ee8e74d912fca8328a353c11d174b44da3f91781},
\release{Eos-0.8B}{4e8776e75a6f0a4bac220e89e82568cb0c66c3c7},
\release{Sol-2B}{bf0248376030f4ef2aca9130f0aa2f64fc255958},
\release{Nox-4B}{bb411db257ca34a99a085764fcf44d3a7c7c2874}, and
\release{Lux-9B}{26559951c5503df8f13c758ae5edb23ca91c9abd}.
The linked repositories provide inference code, configurations, and source attributions. Panel scores can be checked against \texttt{metrics/benchmark.json} in the linked Lux release.

\section{Backbone Dimensions}
\label{app:operators}
Table~\ref{tab:decoder-dimensions} records the deployed causal backbone dimensions. The architecture diagrams in Section~\ref{sec:architecture} show inference computation: dropout is inactive, and kernel-specific layouts are omitted. Fixed source revisions are linked in Appendix~\ref{app:releases}.

\begin{table}[!htbp]
\centering\small
\caption{Decoder backbone dimensions. Q/KV counts describe full-attention query and key/value heads; K/V counts describe Gated DeltaNet key and value heads. Full-attention head width is 256; DeltaNet key and value widths are 128. These are backbone dimensions, distinct from the 256-dimensional decision head.}
\label{tab:decoder-dimensions}
\begin{tabularx}{\textwidth}{lrrrrrX}
\toprule
Model & Layers & Width & FFN & Q/KV & K/V & Mixer schedule \\
\midrule
Eos-0.8B & 24 & 1,024 & 3,584 & 8/2 & 16/16 & 18 GDN + 6 full \\
Sol-2B & 24 & 2,048 & 6,144 & 8/2 & 16/16 & 18 GDN + 6 full \\
Nox-4B & 32 & 2,560 & 9,216 & 16/4 & 16/32 & 24 GDN + 8 full \\
Lux-9B & 32 & 4,096 & 12,288 & 16/4 & 16/32 & 24 GDN + 8 full \\
\bottomrule
\end{tabularx}
\end{table}

\section{Evaluation Details}
\label{app:eval-details}
\paragraph{Panels and aggregation.}
Decisions and Composition contain 880 examples each and macro-average ten task families. Reading contains 480 BoolQ and English/Chinese Belebele examples~\cite{boolq,belebele}, assigning one half of its weight to BoolQ and one quarter to each Belebele language. Inference equally averages four 120-example sources: Cosmos QA, SQuAD answerability, SNLI, and supplied-fact QASC. Transfer uses micro accuracy on the 1,046 answerable examples of the Kev suite. Score accuracy uses the most probable rubric category, not error in the expected ordinal score.

Overall combines the unrounded panel accuracies with the published weights: 30\% Decisions, 25\% Composition, and 15\% each for Reading, Inference, and Transfer. The weights were chosen after observing results and provide a descriptive ranking. Reported differences are also computed before display rounding.

\paragraph{Inference and coverage.}
Untuned Qwen uses fixed answer-letter logits without generated-text parsing. Models retain their native templates and temperatures. Kev uses a BF16 backbone and FP32 head with date injection disabled, differing from its authors' FP32 configuration. Laya truncates some transfer inputs; Jev's server-side input retention is unobservable. Failures count as incorrect when a gold accuracy label exists.

\paragraph{Exposure and uncertainty.}
The external suite is excluded from custom training, checkpoint selection, and temperature fitting, although observed results can inform later research. Paired bootstrap resampling over source groups preserves related examples; Lex's separate comparison resamples its 400 state groups. These intervals quantify example-sampling uncertainty, not benchmark reuse, training-run variation, or unknown upstream pretraining exposure. Differences between architectures and training recipes limit attribution to individual components.

\end{document}